%% file: main.tex
\documentclass[runningheads]{llncs}

\usepackage{eccv}

\usepackage{eccvabbrv}
\input{setup/package}
\input{setup/macros}

\input{setup/symbols}

\input{setup/graphicspath}

\usepackage[pagebackref,breaklinks,colorlinks,citecolor=eccvblue]{hyperref}

\usepackage{orcidlink}

\usepackage[capitalize]{cleveref}
\crefname{section}{Sec.}{Secs.}
\Crefname{section}{Section}{Sections}
\Crefname{table}{Table}{Tables}
\crefname{table}{Tab.}{Tabs.}

\DeclareCaptionLabelFormat{empty}{}
\begin{document}

\title{VideoGigaGAN: Towards Detail-rich Video Super-Resolution} 

\titlerunning{VideoGigaGAN}

\author{Yiran Xu\inst{1,2} \and
Taesung Park\inst{1} \and
Richard Zhang\inst{1} \and
Yang Zhou\inst{1}  \and 
Eli Shechtman\inst{1} \and 
Feng Liu\inst{1} \and 
Jia-Bin Huang\inst{2} \and 
Difan Liu\inst{1} 
}

\authorrunning{Xu et al.}

\institute{Adobe Research \and
University of Maryland, College Park\\
\url{http://videogigagan.github.io}
}




\maketitle
\input{figures/teaser}


\input{0_abstract}
\input{1_introduction}
\input{2_related}
\input{3_method}
\input{4_result}

\input{5_limitation}
\input{figures/limitations}
\input{6_conclusion}

%
%
\bibliographystyle{splncs04}
\bibliography{main}

\end{document}


\definecolor{eccvblue}{rgb}{0.12,0.49,0.85}


\title{
\large{VideoGigaGAN: Towards Detail-rich Video Super-Resolution}
\\ \large{Supplementary Material} 
}


\maketitle

\appendix

This supplementary document includes additional quantitative results with SSIM scores, our network architecture, and training and evaluation details.

\section{Additional quantitative comparison}
We include SSIM scores in Table~\ref{tab:main_baselines_supp}.
Similar to our conclusion in the main paper, our model achieves the lowest LPIPS error, showing a detail-rich result. 

\input{tables/main_baselines_full}

\input{tables/gigagan_config}

\section{Network architecture}
\subsection{GigaGAN upsampler}

We show the configurationss of our GigaGAN upsampler in Table~\ref{tab:supp_model_config}.
For the low-pass filters, we use a kernel of $\frac{1}{16}[1,4,6,4,1]$ before the downsampling. 

\subsection{Flow-guided feature propagation module}
We follow the architecture in BasicVSR++~\cite{chan2022basicvsrpp}. We use SPyNet~\cite{ranjan2017optical} as our flow estimator to reduce memory cost.
For the feature extraction, we use 5 residual blocks. The number of residual blocks for propagation is set to 7.
The kernel size of the deformable convolutional netowrk (DCN) is 3.
We encourage readers to refer to BasicVSR++~\cite{chan2022basicvsrpp} for more details.

\section{Training and evaluation details}
\topic{Datasets.} Following previous works~\cite{chan2021basicvsr,Chan_2022_realbasicvsr}, we use {REDS}~\cite{Nah2019reds4} and {Vimeo-90K}~\cite{xue2019toflow} for training purpose.
For REDS, we use clips 000, 011, 015, 020 of the training set for testing, and clips 000, 001, 006, 017 are used for validation, the rest of the clips are used for training. The ground truth has a resolution of $1280 \times 720$.
For Vimeo-90K, in addition to its official test set Vimeo-90-K, we use UDM10~\cite{PFNL} and Vid4~\cite{liu2013vid4} for testing purpose.
The ground truth has a resolution of $448 \times 256$.

\topic{Degradation.}
We use MMagic's~\cite{mmagic2023} script for degradations - Bicubic (BI) and Blur Downsampling (BD). 
For BD, the ground truth is blurred by a Gaussian filter with $\sigma=1.6$, followed by a $4\times$ subsampling.

\topic{Training settings.} We use Adam optimizer~\cite{kingma2014adam} for training with a fixed learning rate of $5\times10^{-5}$.
During training, we randomly crop a $64 \times 64$ patch from each LR input frames at the same location. We use 10 frames of each video and a batch size of 32 for training.
The batch is distributed into 32 NVIDIA A100 GPUs.
The total number of training iterations for each model is $100,000$.

\topic{Test settings.} During the testing, we use the full-frame of the videos. Particularly, for Vimeo-90K-T, we follow its tradition and only evaluate PSNR, SSIM and LPIPS~\cite{zhang2018unreasonable} on the center frame.

\topic{Metrics.}
We consider two aspects in our evaluation: per-frame quality and temporal consistency. 

For \textbf{per-frame quality}, we use \textbf{PSNR, SSIM, and LPIPS}~\cite{zhang2018unreasonable}. Except for REDS4, we evaluate PSNR and SSIM on y-channel following previous works~\cite{chan2021basicvsr,chan2022basicvsrpp,liu2022learning}.

For \textbf{temporal consistency}, we use warping error $E_{\mathrm{warp}}$~\cite{lai2018learning} and proposed referenced warping error $E^{\mathrm{ref}}_{\mathrm{warp}}$. Please refer to our main paper for the definition of $E^{ref}_{warp}$. We use RAFT~\cite{teed2020raft} as our flow estimator when computing temporal consistency.

\section{More visual results}
We encourage readers to refer to our \href{https://videogigagan.github.io/}{project website} more visual results.

\newpage
\bibliographystyle{ACM-Reference-Format}
\bibliography{main}

%% file: setup/package.tex
\usepackage{subcaption}
\usepackage{float}
\usepackage[justification=raggedright]{caption}	
\usepackage{lscape}                                         
\usepackage{wrapfig}

\usepackage[lined,ruled,linesnumbered]{algorithm2e}
\usepackage{animate} 

\usepackage{booktabs}                   
\usepackage{multirow}

\usepackage{makecell}

\usepackage{paralist}
\usepackage{enumitem}
\setlist[enumerate]{itemsep=0mm}

\usepackage{bm}                          
\usepackage{epsfig}                      
\usepackage{graphicx}                  
\usepackage{mathtools}
\usepackage{amssymb,amsmath}   

\usepackage{units}
\usepackage{color}

\usepackage{comment}

\RequirePackage[hyphens]{url}
\usepackage{url}  
\usepackage{xspace}
\usepackage{xcolor}
\usepackage{setspace}
\usepackage{grfext}
\PrependGraphicsExtensions*{.jpg,.png,.PNG}
\usepackage{pifont}
\usepackage{multirow}
\usepackage{bigstrut}
\usepackage{tabularx}

%% file: setup/macros.tex
\usepackage{soul}
\usepackage{svg}

\usepackage[accsupp]{axessibility} 

\def\eg{e.g.,~}               

\DeclareMathAlphabet{\altmathcal}{OMS}{cmsy}{m}{n}
\DeclareMathAlphabet{\mathbfit}{OT1}{ptm}{bx}{it}

\newlength\paramargin
\newlength\figmargin

\newlength\secmargin
\newlength\figcapmargin
\newlength\tabcapmargin

\newcommand{\mpage}[2]
{
\begin{minipage}{#1\linewidth}\centering
#2
\end{minipage}
}

\newcommand{\topic}[1]
{
\vspace{1mm}\noindent\textbf{#1}
}

\long\def\ignorethis#1{}

\newbox\jsavebox%

\makeatletter
\newcommand{\providelength}[1]{%
  \@ifundefined{\expandafter\@gobble\string#1}
   {
    \typeout{\string\providelength: making new length \string#1}%
    \newlength{#1}%
   }
   {
    \sdaau@checkforlength{#1}%
   }%
}

\newcommand{\sdaau@checkforlength}[1]{%
  \edef\sdaau@temp{\expandafter\sdaau@getfive\meaning#1TTTTT$}%
  \ifx\sdaau@temp\sdaau@skipstring
    \typeout{\string\providelength: \string#1 already a length}%
  \else
    \@latex@error
      {\string#1 illegal in \string\providelength}
      {\string#1 is defined, but not with \string\newlength}%
  \fi
}
\def\sdaau@getfive#1#2#3#4#5#6${#1#2#3#4#5}
\edef\sdaau@skipstring{\string\skip}
\makeatother

%% file: setup/symbols.tex
\def\xi{\mathbf{x}_i}

%% file: setup/graphicspath.tex
\graphicspath{{figures}, {examples}}

%% file: figures/teaser.tex
\begin{center}
\centering

\includegraphics[trim=0 0 0 0, clip,width=\textwidth]{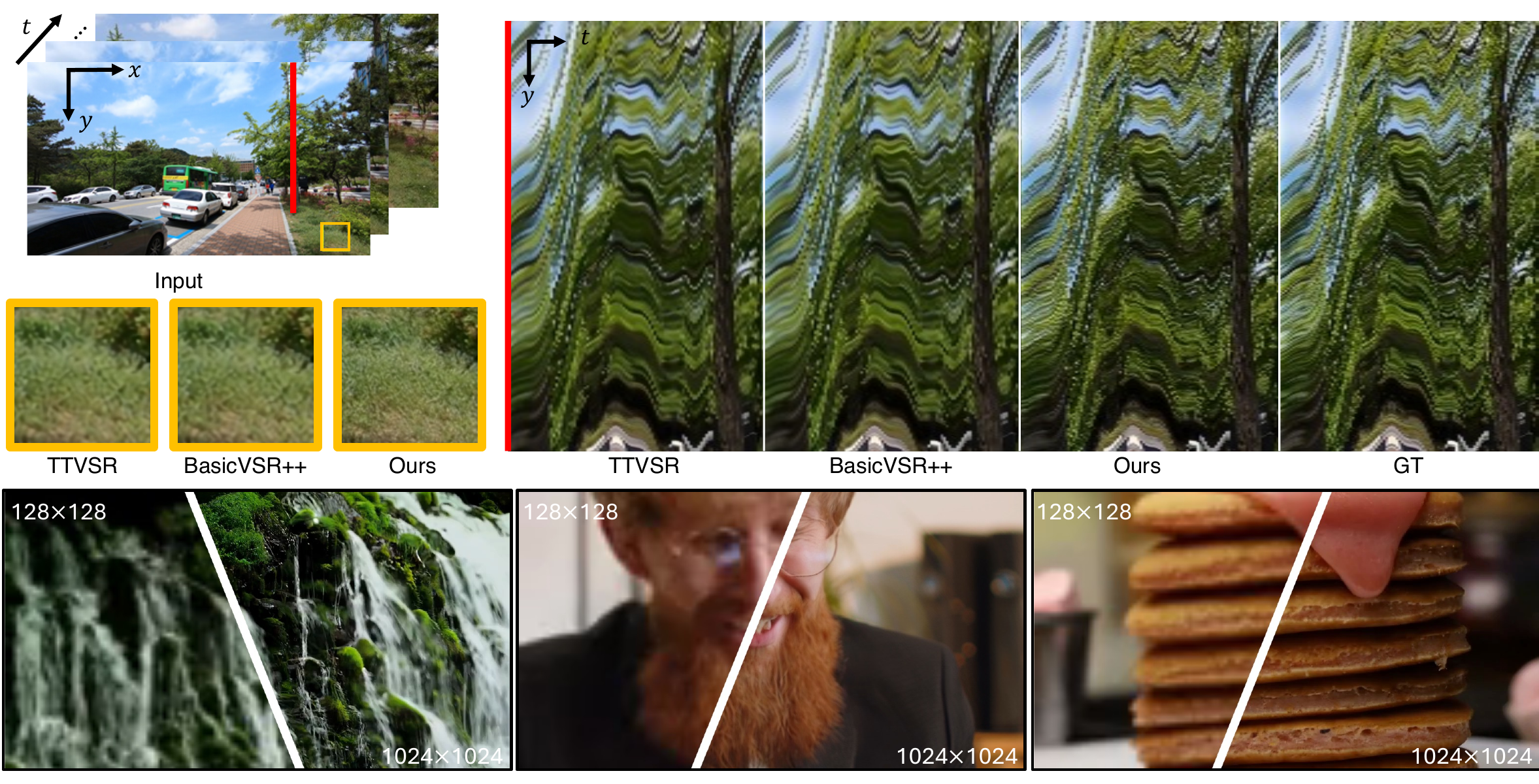}
    \vspace{-5mm}

\captionof{figure}{
We present \textbf{VideoGigaGAN}, a generative video super-resolution model that can upsample videos with high-frequency details while maintaining temporal consistency.
\textit{Top: } we show the comparison of our approach with TTVSR \cite{liu2022learning} and BasicVSR++ \cite{chan2022basicvsrpp}.
Our method produces temporally consistent videos with more fine-grained detailed than previous methods.
\textit{Bottom: } our model can produce high-quality videos with $8 \times$ super-resolution. Please see the video results on our \href{https://videogigagan.github.io/}{project page}.
}
\label{fig:teaser}
\end{center}

%% file: 0_abstract.tex
\begin{abstract}
Video super-resolution (VSR) approaches have shown impressive temporal consistency in upsampled videos. 
However, these approaches tend to generate blurrier results than their image counterparts as they are limited in their generative capability.
This raises a fundamental question: can we extend the success of a generative image upsampler to the VSR task while preserving the temporal consistency?
We introduce VideoGigaGAN, a new generative VSR model that can produce videos with high-frequency details and temporal consistency.
VideoGigaGAN builds upon a large-scale image upsampler -- GigaGAN. 
Simply inflating GigaGAN to a video model by adding temporal modules produces severe temporal flickering.
We identify several key issues and propose techniques that significantly improve the temporal consistency of upsampled videos.
Our experiments show that, unlike previous VSR methods, VideoGigaGAN generates temporally consistent videos with more fine-grained appearance details.
We validate the effectiveness of VideoGigaGAN by comparing it with state-of-the-art VSR models on public datasets and showcasing video results with $8 \times$ super-resolution.

%

\end{abstract}

%% file: 1_introduction.tex
\section{Introduction}
\label{sec:intro}
        
Video super-resolution (VSR) is a classical but challenging task in computer vision and graphics, aiming to recover high-resolution videos from their low-resolution counterparts. 
VSR has two main challenges. 
The first challenge is to maintain temporal consistency across output frames. 
The second challenge is to generate high-frequency details in the upsampled frames.
Previous approaches \cite{isobe2020video, chan2021basicvsr, chan2022basicvsrpp, Chan_2022_realbasicvsr} focus on addressing the first challenge and have shown impressive temporal consistency in upsampled videos. 
However, these approaches often produce blurry results and fail to produce high-frequency appearance details or realistic textures (see Fig.~\ref{fig:motivation}).
An effective VSR model needs to generate plausible new contents not present in the low-resolution input videos.
Current VSR models, however, are limited in their generative capability and unable to hallucinate detailed appearances. 

Generative Adversarial Networks (GANs) \cite{goodfellow2014generative} have shown impressive generative capability on the task of image super-resolution \cite{wang2018esrgan, wang2021realesrgan}. 
These methods can effectively model the distribution of high-resolution images and generate fine-grained details in upsampled images. 
GigaGAN \cite{kang2023gigagan} further increases the generative capability of image super-resolution models by training a large-scale GAN model on billions of images. 
GigaGAN can generate highly detailed textures even for $8 \times$ upsampling tasks. 
However, applying GigaGAN or other GAN-based image super-resolution models to each low-resolution video frame independently leads to severe temporal flickering and aliasing artifacts (see Fig.~\ref{fig:motivation}). 
In this work, we ask -- is it possible to apply GigaGAN for video super-resolution while achieving temporal consistency in upsampled videos?

\input{figures/motivation}

We first experiment with a baseline of inflating the GigaGAN by adding temporal convolutional and attention layers. 
These simple changes alleviate the temporal inconsistency, but the high-frequency details of the upsampled videos are still flickering over time.
As blurrier upsampled videos inherently exhibit better temporal consistency, the capability of GANs to hallucinate high-frequency details contradicts the goal of VSR in producing temporally consistent frames.
We refer to this as the \emph{consistency-quality dilemma} in VSR.
Previous VSR approaches use regression-based networks to trade high-frequency details for better temporal consistency.
In this work, we identify several key issues of applying GigaGAN for VSR and propose techniques to achieve detailed and temporally consistent video super-resolution.
Naively inflating GigaGAN with temporal modules~\cite{ho2022imagen} is not sufficient to produce temporally consistent results with high-quality frames. 
To address this issue, we employ a \textit{recurrent flow-guided feature propagation module} to encourage information aggregation across different frames. 
We also apply \textit{anti-aliasing blocks} in GigaGAN to address the temporal flickering caused by the aliased downsampling operations. 
Furthermore, we introduce an effective method for injecting high-frequency features into the GigaGAN decoder, called \textit{high-frequency (HF) shuttle} . 
The proposed high-frequency shuttle can effectively add fine-grained details to the upsampled videos while mitigating aliasing or temporal flickering.

\paragraph{Contributions.} 
We present VideoGigaGAN, the first large-scale GAN-based model for video super-resolution. 
We recognize the consistency-quality trade-off that has not been well discussed in previous VSR literature.
We introduce the feature propagation module, anti-aliasing blocks and HF shuttle which significantly improve the temporal consistency when applying GigaGAN for VSR.
We show that VideoGigaGAN can upsample videos with much more fine-grained details than state-of-the-art methods evaluated on multiple datasets.
We also show that our model can produce detailed and temporally consistent videos even for challenging $8 \times$ upsampling tasks.

%% file: figures/motivation.tex
\begin{figure*}
\begin{center}
\centering



\begin{minipage}[c]{0.3\textwidth}
    \includegraphics[trim=180 250 550 0, clip, height=1.08\linewidth]{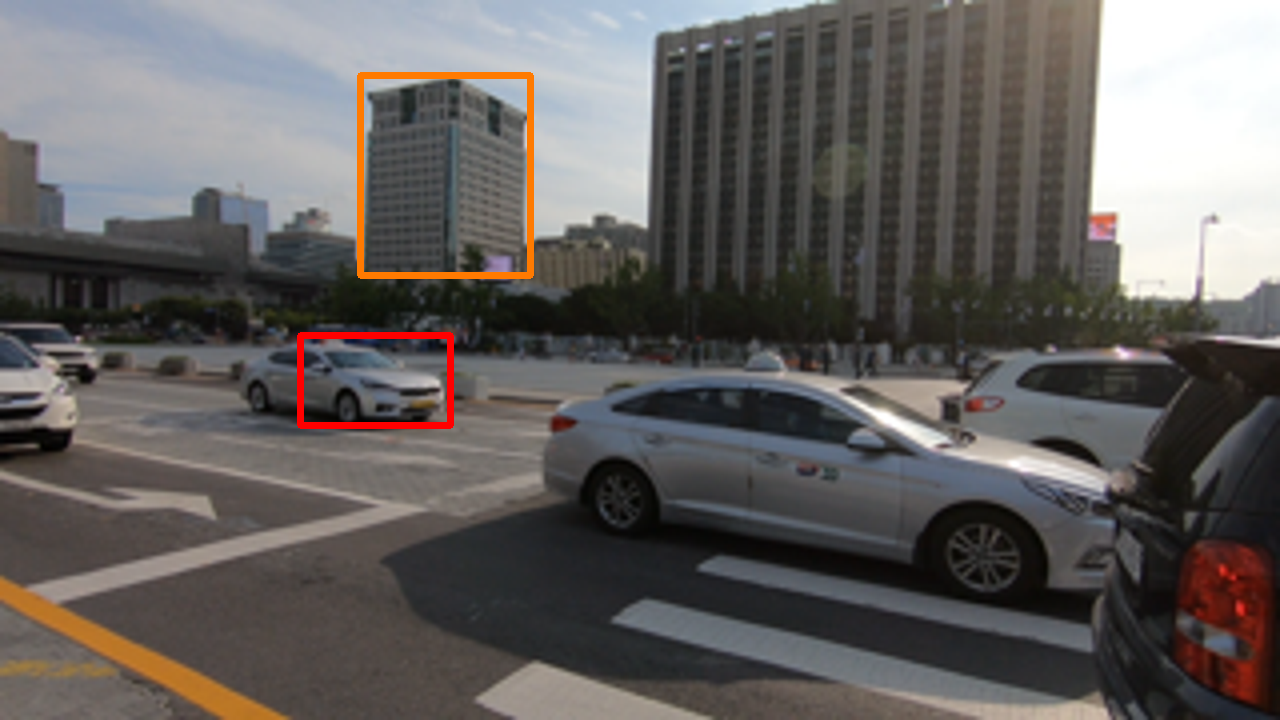}
\end{minipage}
\hfill
\hspace{-5mm}
\begin{minipage}[c]{0.6\textwidth}
    \begin{tabular}{ccc}
        \includegraphics[width=0.3\linewidth]{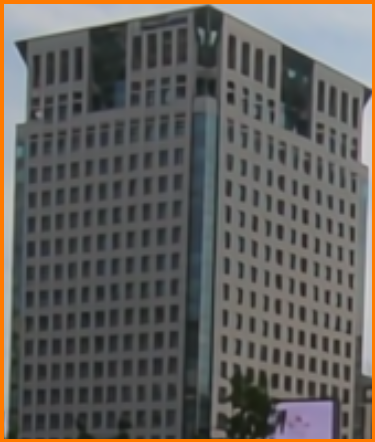} &
        \includegraphics[width=0.3\linewidth]{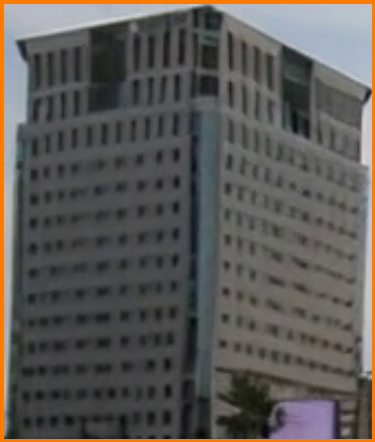} &
        \includegraphics[width=0.3\linewidth]{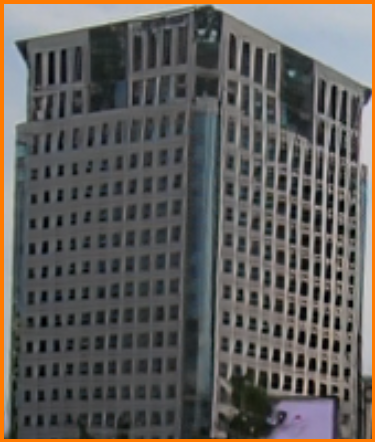} \\
        \includegraphics[width=0.3\linewidth]{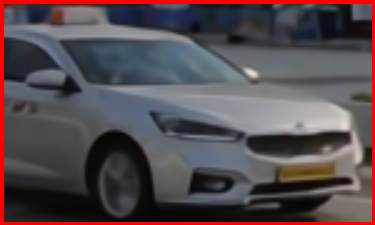} &
        \includegraphics[width=0.3\linewidth]{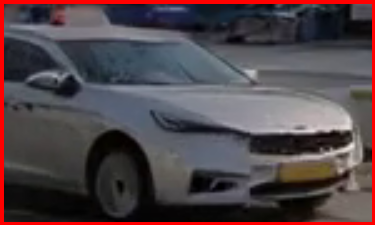} &
        \includegraphics[width=0.3\linewidth]{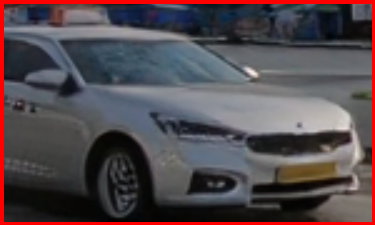}
    \end{tabular}
\end{minipage}

\mpage{0.35}{\small{Input}}\hfill
\mpage{0.25}{\small{BasicVSR++}}\hfill
\mpage{0.12}{\small{GigaGAN}}\hfill
\mpage{0.22}{\small{Ours}}

\caption{
\textbf{Limitations of previous methods.} Previous VSR approaches such as BasicVSR++~\cite{chan2022basicvsrpp} suffer from lack of details, as seen from the \textcolor{red}{car} example. 
Image GigaGAN 
produces sharper results with richer details, but it generates videos with temporal flickering and artifacts like aliasing (see \textcolor{orange}{building}). 
Our VideoGigaGAN can produce video results with both high-frequency details and temporal consistency while artifacts like aliasing are significantly mitigated.
}
\label{fig:motivation}
\end{center}
\end{figure*}

%% file: 2_related.tex
\section{Related Work}
\label{sec:related}

\topic{Video Super-Resolution.}
Significant work has been invested in video super-resolution, using sliding-window approaches~\cite{caballero2017real, tao2017detail, xue2019toflow, wang2019deformable, li2020mucan, tian2020tdan} and recurrent networks~\cite{huang2015bidirectional, huang2017video, sajjadi2018frame, isobe2020video, liang2022recurrent, li2023multi, shi2022rethinking, liang2022vrt}.
BasicVSR~\cite{chan2021basicvsr} summarizes the common VSR approaches into a unified pipeline. 
It proposes an effective baseline using optical flow for temporal alignment and bidirectional recurrent networks for feature propagation.
BasicVSR++~\cite{chan2022basicvsrpp} redesigns BasicVSR by introducing second-order grid propagation and flow-guided deformable alignment.	
To improve the generalizability on real-world low-resolution videos, methods like RealBasicVSR \cite{Chan_2022_realbasicvsr} and FastRealVSR \cite{xie2023mitigating} use diverse degradations as data augmentation during training.
While these approaches can produce temporally consistent upsampled videos, they are often trained with simple regression objectives and lack the generative capability, which leads to unrealistic textures and overly blurry results.
Unlike previous VSR approaches, we propose a GAN-based VSR model to generate high-frequency details while maintaining temporal consistency in the upsampled videos.

\topic{GAN-based Image Super-Resolution.}
SRGAN~\cite{ledig2017photo} is a seminal image super-resolution work that uses a GAN framework to model the manifold of high-resolution images. 
ESRGAN \cite{wang2018esrgan} further enhances the visual quality of upsampled images by improving the architecture and loss of SRGAN.
Real-ESRGAN \cite{wang2021realesrgan} extends ESRGAN to restore general real-world low-resolution images.
While these methods can produce impressive results, they are still limited in model capacity and unsuitable for large upsampling factors.
To scale up the model capacity of GANs, GigaGAN \cite{kang2023gigagan} introduces filter bank and attention layers to StyleGAN2 \cite{karras2020stylegan2} and trains the model on billions of images.
Even for $8 \times$ image super-resolution tasks, GigaGAN can effectively generate new content not present in the low-resolution image and produce realistic textures and fine-grained details.

\topic{Generative Video Models.}
Many video generation works are based on the VAEs~\cite{babaeizadeh2017stochastic,lee2018stochastic,yan2021videogpt}, GANs~\cite{skorokhodov2022stylegan,ge2022long,zhang2022towards}, and autoregressive models \cite{weissenborn2019scaling}.
LongVideoGAN \cite{brooks2022generating} introduces a sliding-window approach for video super-resolution, but it is restricted to datasets with limited diversity.
Recently, diffusion models have shown diverse and high-quality results in video generation tasks \cite{ho2022video, ge2023preserve, blattmann2023align, girdhar2023emu, blattmann2023stable}.
Imagen Video \cite{ho2022imagen} proposes pixel diffusion models for video super-resolution.
Concurrent work Upscale-A-Video \cite{zhou2023upscale} adds temporal modules to a latent diffusion image upsampler \cite{rombach2022sd} and finetunes it as a video super-resolution model.
Unlike diffusion-based video super-resolution models that require iterative denoising processes, our VideoGigaGAN can generate outputs in a \emph{single feedforward pass} with faster inference speed. 



%% file: 3_method.tex
\def\D{\altmathcal{D}}
\def\I{\altmathcal{I}}
\def\O{\altmathcal{O}}
\def\res{\altmathcal{R}}

\def\b{\mathbfit{b}}
\def\c{\mathbfit{c}}
\def\d{\mathbfit{d}}
\def\o{\mathbfit{o}}
\def\p{\mathbfit{p}}
\def\t{\mathbfit{t}}
\def\x{\mathbfit{x}}
\def\z{\mathbfit{z}}

\def\K{\mathbfit{K}}
\def\R{\mathbfit{R}}

\def\ang{\phi}
\def\dehom{\mu}
\def\proj{\pi}
\def\sigmoid{S}
\def\vis{\nu}
\def\r{\mathbfit{r}}

\def\bp{(\p\!)} 
\def\bt{(t\!)} 
\def\bx{(\x\neg)} 

\def\ok{\o_{\neg k}}
\def\tk{\t_{\neg k}}
\def\wk{w_{\neg k}}
\def\xi{\x_{\neg i}}
\def\zk{\z_{\neg k}}
\def\Kk{\K_{\neg k}}
\def\Rk{\R_{\neg k}}

\def\ng{\hspace{-0.1mm}}
\def\neg{\hspace{-0.2mm}}
\def\pos{\hspace{0.2mm}}

\makeatletter
\newcommand*\MY@rightharpoonupfill@{%
    \arrowfill@\relbar\relbar\rightharpoonup
}
\newcommand*\overrightharpoon{%
    \mathpalette{\overarrow@\MY@rightharpoonupfill@}%
}
\makeatother

\newlength{\depthofsumsign}
\setlength{\depthofsumsign}{\depthof{$\sum$}}
\newcommand{\nsum}[1][1.4]{
    \mathop{%
        \raisebox
            {-#1\depthofsumsign+1\depthofsumsign}
            {\scalebox
                {#1}
                {$\displaystyle\sum$}%
            }
    }
}

\input{figures/method_overview}
\section{Method}
\label{sec:method}

Our VSR model $\mathcal{G}$ upsamples a low-resolution (LR) video $\mathbf{v} \in \mathbb{R}^{T \times h \times w \times 3}$ to a high-resolution (HR) video $\mathbf{V} = \mathcal{G}(\mathbf{v})$, where $\mathbf{V} \in \mathbb{R}^{T \times H \times W \times 3}$, with an upsampling scale factor $\alpha$ such that $H = \alpha h,\  W = \alpha w$.
We aim to generate HR videos with both high-frequency appearance details and temporal consistency.

We present the overview of our VSR model, \textbf{VideoGigaGAN}, in Fig.~\ref{fig:method_overview}.
We start with the large-scale GAN-based image upsampler -- GigaGAN~\cite{kang2023gigagan} (Section~\ref{sec:pre_gigagan}). 
We first inflate the 2D image GigaGAN upsampler to a 3D video GigaGAN upsampler by adding temporal convolutional and attention layers (Section~\ref{sec:method:inflation}). 
However, as shown in our experiments, the inflated GigaGAN still produces results with severe temporal flickering and artifacts, likely due to the limited spatial window size of the temporal attention. 
To this end, we introduce flow-guided feature propagation (Section~\ref{sec:method:propagation}) to the inflated GigaGAN to better align the features of different frames based on flow information. 
We also pay special attention to anti-aliasing (Section~\ref{sec:antialias}) to further mitigate the temporal flickering caused by the downsampling blocks in the GigaGAN encoder, while maintaining the high-frequency details by directly shuttling the HF features to the decoder blocks (Section~\ref{sec:HF}).
Our experimental results validate the importance of these model design choices.

\subsection{Preliminaries: Image GigaGAN upsampler}\label{sec:pre_gigagan}
Our VideoGigaGAN builds upon the GigaGAN image upsampler \cite{kang2023gigagan}. 
GigaGAN scales up the StyleGAN2~\cite{karras2020stylegan2} architecture using several key components, including adaptive kernel selection for convolutions and self-attention layers.
The GigaGAN image upsampler has an asymmetric U-Net architecture consisting of $3$ downsampling blocks $\{E_i\}$ and $3 + k$ upsampling decoder blocks $\{D_i\}$. 
\begin{equation}
    \begin{split}
        \mathbf{X} &= \mathcal{G}(\mathbf{x}, \mathbf{z})  = D(E(\mathbf{x}, \mathbf{z}), \mathbf{z}) \\
        &= \underbrace{D_{k+2} \circ \cdots D_{3}}_{\uparrow \times 2^k} \circ \underbrace{D_{2} \circ D_{1} \circ D_{0}}_{\uparrow \times 8} \circ \underbrace{E_{2} \circ E_{1} \circ E_{0} (\mathbf{x}, \mathbf{z})}_{\downarrow \times 8}  \,.
    \end{split}
    \label{eqn:gigagan}
\end{equation}
This GigaGAN upsampler is able to upsample an input image by $2^{k}$.
Both encoder $E$ and decoder $D$ blocks utilize random spatial noise $\mathbf{z}$ as a source of stochasticity. 
The decoder $D$ contains spatial self-attention layers.  
The encoder and decoder block at same resolution are connected by skip connections.


\subsection{Inflation with temporal modules}
\label{sec:method:inflation}
To adapt a pretrained 2D image model for video tasks, a common approach is to inflate 2D spatial modules into 3D temporal ones~\cite{zhou2023upscale,blattmann2023align,ge2023preserve,wu2023tune,yang2023rerender,ho2022imagen}. 
To reduce the memory cost, instead of directly using 3D convolutional layers in each block, our temporal module uses a 1D temporal convolution layer that only operates on the temporal dimension of kernel size 3, followed by a temporal self-attention layer with no spatial receptive field. 
Both 1D temporal convolution and temporal self-attention are inserted after the spatial self-attention with residual connection~\cite{ho2022imagen}. 
In summary, at each block $D_{i}$, we first process the features of individual video frames using the spatial self-attention layer and then jointly processed by our temporal module.
Through our experiment, we find adding temporal modules to the decoder $D$ of the generator $\mathcal{G}$ is sufficient to improve video consistency. 
We also inflate the discriminator $\mathcal{D}$ with comparable temporal modules.

We follow \cite{zhang2023adding} to initialize both temporal convolutions and temporal self-attention layers with zero weights, such that $\mathcal{G}$ and $\mathcal{D}$ still perform the same as an image upsampler at the beginning of the training, leading to a smoother transition to a video upsampler.

\subsection{Flow-guided feature propagation}
\label{sec:method:propagation}
The temporal modules alone are insufficient to ensure temporal consistency, mainly due to the high memory cost of the 3D layers. 
For input videos with long sequences of frames, one could partition the video into small, non-overlapping chunks and apply temporal attention. 
However, this leads to temporal flickering between different chunks. 
Even within each chunk, the spatial window size of the temporal attention is limited, meaning a large motion (i.e., exceeding the receptive field) cannot be modeled by the attention module (see Fig.~\ref{fig:ablation}). 

To address these issues, we augment the input image with features aligned by optical flow. 
Specifically, we introduce a recurrent flow-guided feature propagation module (see Fig.~\ref{fig:method_overview}) prior to the inflated GigaGAN, inspired by BasicVSR++ \cite{chan2022basicvsrpp}. 
Instead of directly using the LR video as input to the inflated GigaGAN, we use the temporal-aware features produced by the flow-guided propagation module. 
It comprises a bi-directional recurrent neural network (RNN)~\cite{chan2021basicvsr,chan2022basicvsrpp} and an image backward warping layer.
We initially employ the optical flow estimator to predict bi-directional optical flow maps from the input LR video.
Subsequently, these maps and the original frame pixels are fed into the RNN to learn temporal-aware features. 
Finally, these features are explicitly warped using the backward warping layer, guided by the pre-computed optical flows, before being fed into the later inflated GigaGAN blocks. 
The flow-guided propagation module can effectively handle large motion and produce better temporal consistency in output videos, as demonstrated in Fig~\ref{fig:ablation}. 

During training, we jointly train the flow-guided feature propagation module and the inflated GigaGAN model. 
At inference time, given an input LR video with an arbitrary number of frames, we first generate frame features using the flow-guided propagation module. 
We then partition the frame features into non-overlapping chunks and independently apply the inflated GigaGAN on each chunk.
Since the features inside each chuck are \emph{aware} of the other chunks, thanks to the flow-guided propagation module, the temporal consistency between consecutive chunks is preserved well.

\subsection{Anti-aliasing blocks}
\label{sec:antialias}
With both temporal and feature propagation modules enabled, our VSR model can process longer videos and produce results with better temporal consistency.
However, the high-resolution frames remain flickering in areas with high-frequency details (for example, the windows in the building in Fig.~\ref{fig:motivation}).
We identify that the downsampling operations in the GigaGAN encoder contribute to the flickering of those regions.
The high-frequency components in the input can easily alias into lower frequencies due to the downsampling rate not meeting the classical sampling criterion \cite{nyquist1928certain}. 
The aliasing of pixels manifests as temporal flickering in video super-resolution. 
Previous VSR approaches often use regression-based objectives, which tend to remove high-frequency details.
Consequently, these methods produce output videos free of aliasing.
However, in our GAN-based VSR framework, the GAN training objectives favor the hallucination of high-frequency details, making aliasing a more severe problem.

In the GigaGAN upsampler, the downsampling operation in the encoder is achieved by strided convolutions with a stride of 2.
To address the aliasing issue in our output video, we apply BlurPool layers to replace all the strided convolution layers in the upsampler encoder inspired by~\cite{zhang2019shiftinvar}.
More specifically, during downsampling, instead of simply using a strided convolution, we use convolution with a stride of $1$, followed by a low-pass filter and a subsampling operation. 
We show the anti-aliasing blocks in Fig.~\ref{fig:method_overview}. 
Our experiments show that the anti-aliasing downsampling blocks perform significantly better than naive strided convolutions in preserving temporal consistency for high-frequency details.
We also experimented with StyleGAN3 blocks for anti-aliasing upsampling \cite{karras2021alias}. 
The temporal flickering is mitigated, but we observed a notable drop in frame quality. 

\subsection{High-frequency shuttle}
\label{sec:HF}
With the newly introduced components, the temporal flicker in our results is significantly suppressed. 
However, as shown in Fig.~\ref{fig:ablation}, adding the flow-guided propagation module (Section~\ref{sec:method:propagation}) leads to a blurrier output.
Anti-aliasing blocks (Section~\ref{sec:antialias}) make the results even blurrier.
We still need the high-frequency information in the GigaGAN features to compensate for the loss of high-frequency details. 
However, as discussed in Section~\ref{sec:antialias}, the traditional flow of high-frequency information in GigaGAN leads to aliased output.

We present a simple yet effective approach to address the conflict of high-frequency details and temporal consistency, called \emph{high-frequency shuttle} (HF shuttle). 
To guide where the high-frequency details should be inserted, the HF shuttle leverages the skip connections in the U-Net and uses a pyramid-like representation for the feature maps in the encoder. 
More specifically, at the feature resolution level $i$, we decompose the feature map $f_{i}$ into low-frequency (LF) feature and high-frequency (HF) components. 
The LF feature map $f^{LF}_{i}$ is obtained via the low-pass filter mentioned in Section~\ref{sec:antialias}, while the HF feature map is computed from the residual as $f^{HF}_{i}=f_{i} - f^{LF}_{i}$. 
The HF feature map $f^{HF}_{i}$ containing high-frequency details are injected through the skip connection to the decoder (Fig.~\ref{fig:method_overview}). 
Our experiments show that the high-frequency shuttle can effectively add fine-grained details to the upsampled videos while mitigating issues such as aliasing or temporal flickering.

\subsection{Loss functions}
We use stardard, non-saturating GAN loss~\cite{gulrajani2017improved}, R1 regularization~\cite{mescheder2018training}, LPIPS~\cite{zhang2018unreasonable} and Charbonnier loss~\cite{charbonnier1994two} during the training.
\begin{equation}\label{eqn:loss}
    \begin{split}
        \mathcal{L}(\mathbf{X}_t, \mathbf{x}_t) & = 
        \mu_{GAN} \mathcal{L}_{GAN}(\mathcal{G}(\mathbf{x}_t), \mathcal{D}(\mathcal{G}(\mathbf{x}_t))) 
    + \mu_{R1} \mathcal{L}_{R1}(\mathcal{D}(\mathbf{X}_t))\\
    &+ \mu_{LPIPS}\mathcal{L}_{LPIPS}(\mathbf{X}_t, \mathbf{x}_t) + \mu_{Char}\mathcal{L}_{Char}(\mathbf{X}_t, \mathbf{x}_t)  \,,
    \end{split}
\end{equation}
where Charbonnier loss is a smoothed version of pixelwise $\ell_1$ loss, $\mu_{GAN}, \mu_{R1}$, $\mu_{LPIPS}, \mu_{Char}$ are the scales of different loss functions. $\mathbf{x}_t$ is one of the LR input frames, $\mathbf{X}_t$ is the corresponding ground-truth HR frame. 
We average the loss over all the frames in a video clip during the training.

%% file: figures/method_overview.tex
\definecolor{color_temporal}{HTML}{E5F0DA}
\definecolor{color_flow}{HTML}{7F7F7F}
\definecolor{color_blurpool}{HTML}{B8C7E4}
\definecolor{color_shuttlehf}{HTML}{F1CCB1}

\begin{figure*}[t!]
\begin{center}
\centering
\includegraphics[trim=0 0 0 0, clip,width=0.9\textwidth]{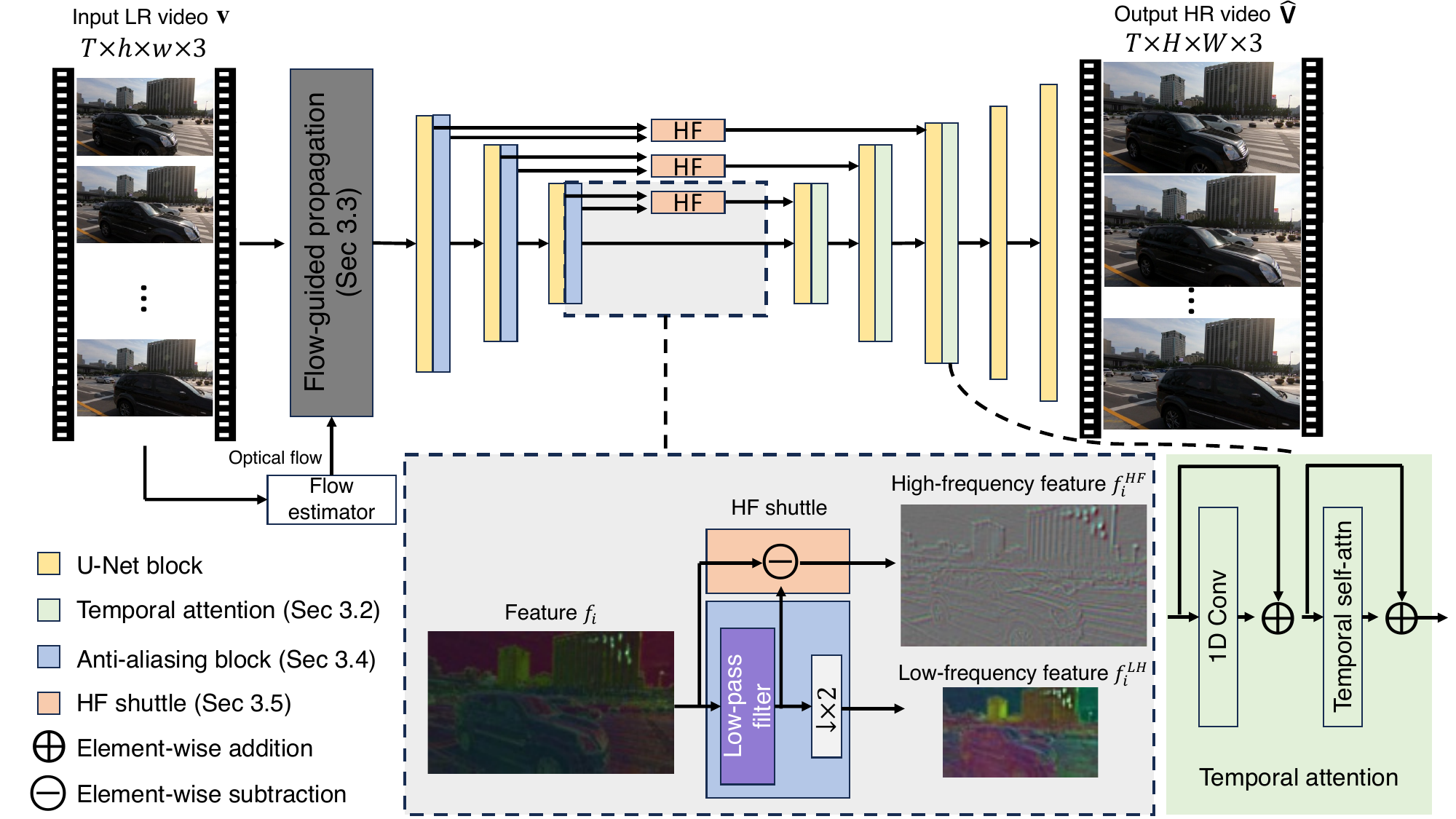}
\caption{
\textbf{Overview of our method} for $4\times$ upsampling. 
Our Video Super-Resolution (VSR) model is built upon the asymmetric U-Net architecture of the image GigaGAN upsampler~\cite{kang2023gigagan}. To enforce temporal consistency, we first inflate the image upsampler into a video upsampler by adding \colorbox{color_temporal}{{temporal attention}} layers into the decoder blocks. We also enhance consistency by incorporating the features from the \colorbox{color_flow}{flow-guided propagation} module. To suppress aliasing artifacts, we use \colorbox{color_blurpool}{Anti-aliasing block} in the downsampling layers of the encoder. Lastly, we directly \colorbox{color_shuttlehf}{{shuttle the high frequency features}} via skip connection to the decoder layers to compensate for the loss of details in the BlurPool process. 
}
\label{fig:method_overview}
\end{center}
\end{figure*}

%% file: 4_result.tex
\section{Experimental Results}
\label{sec:result}
\input{figures/qualitative_baselines}
\input{figures/ablation}
\subsection{Setup}
\label{sec:setup}
\topic{Datasets.} 
We strictly follow two widely used training sets from previous VSR works
~\cite{chan2022basicvsrpp,liu2022learning,chan2021basicvsr}: \textbf{REDS}~\cite{Nah2019reds4} and \textbf{Vimeo-90K}~\cite{xue2019toflow}. 
The REDS dataset contains 300 video sequences. 
Each sequence consists of 100 frames with a resolution of $1280\times720$.
We use REDS4 as our test set and REDSval4 as our validation set; the rest of the sequences are used for training.
The Vimeo-90K contains $64,612$ sequences for training and $7,824$ for testing (known as Vimeo-90K-T). 
Each sequence contains seven frames with a resolution of $448\times256$.
Following previous works~\cite{chan2021basicvsr, chan2022basicvsrpp}, we compute the metrics only on the center frame of each sequence. 
In addition to the official test set Vimeo-90K-T, we also evaluate the model on Vid4~\cite{liu2013vid4} and UDM10~\cite{PFNL}, with different degradation algorithms (Bicubic Downsampling -- BI and Blur Downsampling -- BD).
We follow MMagic~\cite{mmagic2023} to perform degradation algorithms.
All data are $4\times$ downsampled to generate LR frames following standard evaluation protocols~\cite{chan2021basicvsr, chan2022basicvsrpp}.

\topic{Evaluation metrics.} 
We are interested in two aspects of our evaluation: \emph{per-frame quality} and \emph{temporal consistency}. 
For per-frame quality, we use \textbf{PSNR, SSIM,} and \textbf{LPIPS}~\cite{zhang2018unreasonable}.
We report SSIM scores in the \href{https://videogigagan.github.io/assets/supp.pdf}{supplementary material}.
For temporal consistency, the warping error $E_{\mathrm{warp}}$~\cite{lai2018learning} is commonly used. 
\begin{equation}
    E_{\mathrm{warp}}(\hat{\mathbf{X}}_t,\hat{\mathbf{X}}_{t+1}) = \frac{1}{\sum M^{i}_t} \sum M^{i}_t ||\hat{\mathbf{X}}^{i}_{t}, W(\hat{\mathbf{X}}^{i}_{t+1}, \mathcal{F}_{t \rightarrow t+1}) ||^2_2 \,,
\end{equation}
where $(\hat{\mathbf{X}}_t,\hat{\mathbf{X}}_{t+1})$ are \textbf{generated} frames at time $t$ and $t+1$, $i$ is the index of the $i$-th pixel, and $W(\cdot)$ is the warping function, $\mathcal{F}_{t \rightarrow t+1}$ is the forward flow estimated from the generated frames $(\hat{\mathbf{X}}_t,\hat{\mathbf{X}}_{t+1})$ using RAFT ~\cite{teed2020raft}, and $M_t \in \{0, 1\}$ is a non-occlusion mask indicating non-occluded pixels~\cite{ruder2016artistic}.
However, as reported in Table~\ref{tab:trade_off}, previous baselines such as BasicVSR++ or even simple bicubic upsampling achieve lower $E_{\mathrm{warp}}$ than ground truth high-resolution video since $E_{\mathrm{warp}}$ favors over-smoothed results. 
Consider an extreme algorithm where all the generated frames are entirely black. $E_{\mathrm{warp}}$ computes the warping errors by warping the generated frames. The warping error for this algorithm is $\mathbf{0}$ since the generated frames are over-smoothed (in this extreme case, all black).
Therefore, instead of warping the generated frames, we propose to warp the ground-truth frames using the flow computed on the generated frames. 
We refer to this new warping error as \textbf{referenced warping error} $E^{\mathrm{ref}}_{\mathrm{warp}}$. The referenced warping error between two frames is
\begin{equation}
    E^{ref}_{\mathrm{warp}}({\mathbf{X}}_t,{\mathbf{X}}_{t+1}) = \frac{1}{\sum M^{i}_t} \sum M^{i}_t ||\mathbf{X}^{i}_t, W(\mathbf{X}^{i}_{t+1}, \mathcal{F}_{t \rightarrow t+1})) ||^2_2 \,,
\end{equation}
where $({\mathbf{X}}_t,{\mathbf{X}}_{t+1})$ are ground-truth frames at time $t$ and $t+1$,  $\mathcal{F}_{t \rightarrow t+1}$ is the forward flow estimated from the \textbf{generated} frames $(\hat{\mathbf{X}}_t,\hat{\mathbf{X}}_{t+1})$ using RAFT ~\cite{teed2020raft}.

\topic{Hyperparameters.} We use a pretrained $4\times$ GigaGAN image upsampler as our base model. It contains three downsampling blocks in the encoder and five upsampling blocks in the decoder. 
The spatial self-attention layers are only used in the first block of the decoder for memory efficiency. 
For the flow network, we use a lightweight SpyNet~\cite{ranjan2017optical}.
For the low-pass filters, we use a kernel of $\frac{1}{16}[1,4,6,4,1]$ before the downsampling. 
We set $\mu_{GAN}=0.05$, $\mu_{R1}=0.2048$, $\mu_{LPIPS}=5$, $\mu_{Char}=10$ in Eqn.~\ref{eqn:loss}.
During training, we randomly crop a $64 \times 64$ patch from each LR input frame at the same location. We use 10 frames of each video and a batch size of 32 for training. 
The batch is distributed into 32 NVIDIA A100 GPUs.
We use a fixed learning rate of $5\times10^{-5}$ for both generator and discriminator.
The total number of training iterations is $100,000$.

\subsection{Ablation study}\label{sec:ablation}
To demonstrate the effect of each proposed component, we progressively add them one by one and evaluate them on the REDS4 dataset~\cite{Nah2019reds4}.
We report the quantitative results in Table~\ref{tab:ablation}.
We also present a qualitative comparison in Fig.~\ref{fig:ablation}.
We see that the \textbf{flow-guided feature propagation} brings a large LPIPS and $E^{ref}_{\mathrm{warp}}$ improvement compared to the \textbf{temporal attention}. 
This demonstrates the effectiveness of the feature propagation contributing to the temporal consistency.
By further introducing BlurPool as the \textbf{anti-aliasing} block, the model has a warping error drop but an LPIPS loss increase (also shown in Fig.~\ref{fig:ablation}).
Finally, by using \textbf{HF shuttle}, we can bring the LPIPS back with a slight loss of temporal consistency. Though it is not reflected on the number clearly, we observed that the sharpness of the frame improves significantly with the HF shuttle (see in the x-t slice plot in Fig.~\ref{fig:ablation}). 
We strongly encourage the readers to watch the videos in the \href{https://videogigagan.github.io}{project website}.

\input{tables/ablation}

\input{figures/quality_vs_temp}
\subsection{Comparison with previous models}\label{sec:comp_sotas}

\input{tables/main_baselines}

We conduct extensive experiments by comparing with 9 models including BasicVSR++ \cite{chan2022basicvsrpp} and TTVSR \cite{liu2022learning}.
At this point we cannot include
Upscale-A-Video \cite{zhou2023upscale} since there is no available code.
We report the quantitative comparison of the per-frame quality in Table~\ref{tab:main_baselines}.  
We show the comparison of temporal consistency for 6 of them in Table~\ref{tab:trade_off}.
Additionally, we provide qualitative comparisons in Fig.~\ref{fig:qual_comparison}. 


\topic{Per-frame quality.} As shown in Table~\ref{tab:main_baselines}, our LPIPS outperforms all the other models by a large margin while showing a poorer performance of PSNR and SSIM (for SSIM, please refer to \href{https://videogigagan.github.io/assets/supp.pdf}{supplementary material}).
We observe that PSNR and SSIM do not align well with human perception and favor blurry results, as also reported in the literature~\cite{kang2023gigagan,rombach2022sd, saharia2022image}.
Thus we consider LPIPS~\cite{zhang2018unreasonable} as our core metric to evaluate per-frame quality as it is closer to the human perception.
In Fig.~\ref{fig:qual_comparison}, it is noticeable that our model produces results with the most fine-grained details. Previous approaches tend to predict blurry results with a critical loss of details.

\topic{Temporal consistency.} 
As observed in previous works~\cite{lai2018learning}, the widely used warping error metric favors a more blurry video. 
This is also illustrated in the Table~\ref{tab:trade_off}. The simple bicubic upsampling method achieves the best performance for the commonly used warping error, which is much better than the GT warping error.
We proposed the referenced warping error (\textbf{RWE}) in Section \ref{sec:setup} to address the issue of warping error favoring blurry results.
In terms of the referenced warping error, our method is slightly worse than previous methods ($0.05 \times 10^{-3}$ compared to BasicVSR++~\cite{chan2022basicvsrpp}).
The newly proposed \textbf{RWE} is more suitable for evaluating the temporal consistency of upsampled videos. However, it is still biased towards more blurry results as seen in Table~\ref{tab:trade_off} (several methods, including BasicVSR, BasicVSR++, and TTVSR, are still better than the ground truth high-resolution videos). 
We leave a better metric of VSR temporal consistency for future works.



\subsection{Analysis of the trade-off between temporal consistency and frame fidelity}
\begin{figure}
    \centering
    \includegraphics[width=0.75\textwidth]{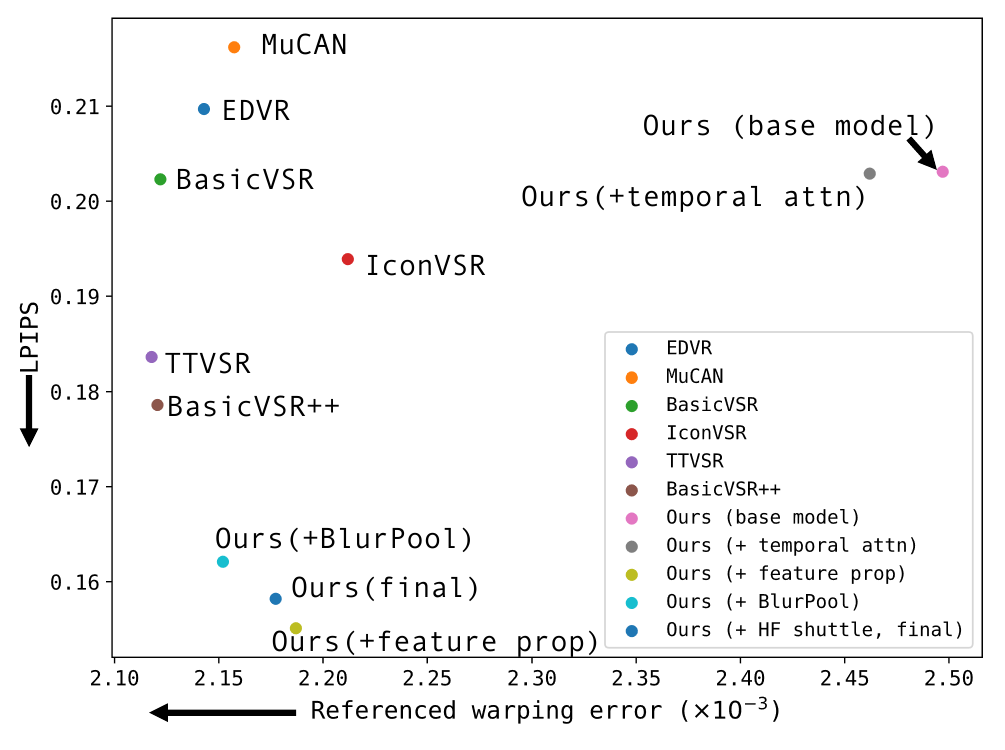}
    \caption{\textbf{Trade-off between per-frame quality (LPIPS$\downarrow$) and temporal consistency (RWE$\downarrow$).} Our final model achieves a good balance between the temporal consistency and per-frame quality.}
    \label{fig:temp_comparison}
    \vspace{-1mm}
\end{figure}

To better understand the trade-off between the temporal consistency and per-frame quality, we include a visualization in Fig.~\ref{fig:temp_comparison}. We can see that the previous VSR approaches focus on achieving better temporal consistency, but this comes with a sacrifice of per-frame quality (also see the qualitative comparisons in Fig.~\ref{fig:qual_comparison}).
Unlike previous VSR approaches, our final model - VideoGigaGAN, achieves a good balance between temporal consistency and per-frame quality. Compared to the base model GigaGAN, our proposed components significantly improve both the temporal consistency and per-frame quality by a large margin. 

\input{tables/modelsize_compute_comp}

\subsection{Model sizes and runtimes}
We show the model sizes and runtimes for different models in Table~\ref{tab:modelsize_runtime}. 
Our model has a large size for its generative capacity,
and still has a competitive inference speed compared to previous feedforward VSR methods. 
Unlike diffusion-based video super-resolution models \cite{ho2022imagen,zhou2023upscale} that require iterative denoising processes, our VideoGigaGAN can generate outputs in a \emph{single feedforward pass} with much faster inference speed. 
We also experimented with scaling previous feed-forward models such as BasicVSR++ \cite{chan2022basicvsrpp}. However, previous VSR models do not have good scalability and show unstable training when scaling up as also discussed in \cite{kang2023gigagan}. 

\subsection{$8\times$ video upsampling}
Our model is capable for 8x video upsampling with both good temporal consistency and per-frame quality with rich details.
We encourage readers to visit our \href{https://videogigagan.github.io/}{project website} for more results.


%% file: figures/qualitative_baselines.tex
\begin{figure*}
    \mpage{0.15}{\scriptsize{Input}}\hfill
    \mpage{0.15}{\scriptsize{BasicVSR~\cite{chan2021basicvsr}}}\hfill
    \mpage{0.15}{\scriptsize{TTVSR~\cite{liu2022learning}}}\hfill
    \mpage{0.15}{\scriptsize{BasicVSR++~\cite{chan2022basicvsrpp}}}\hfill
    \mpage{0.15}{\scriptsize{Ours}}\hfill
    \mpage{0.15}{\scriptsize{GT}}    
    \begin{subfigure}[t]{0.15\textwidth}
        \centering
        \frame{\includegraphics[width=\textwidth]{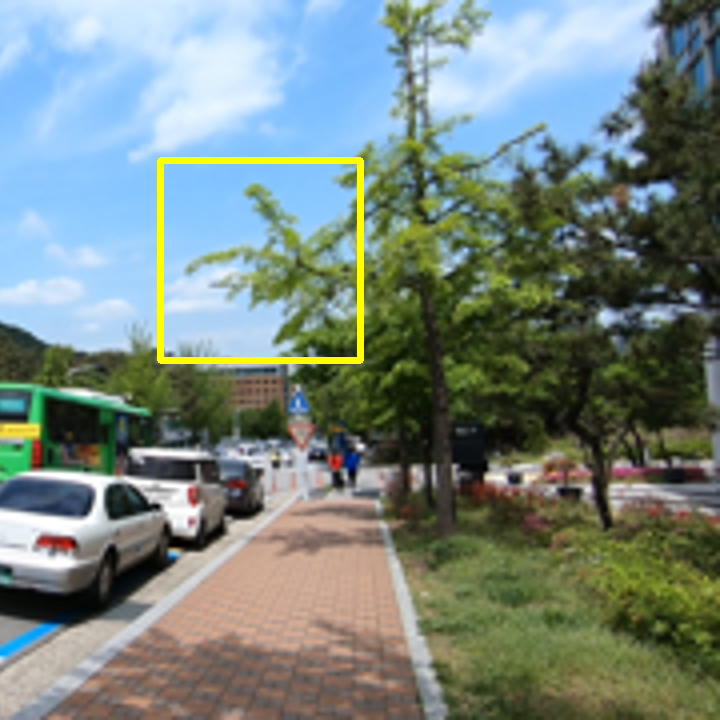}}
    \end{subfigure}
    \hfill
    \begin{subfigure}[t]{0.15\textwidth}
        \centering
        \frame{\includegraphics[width=\textwidth]{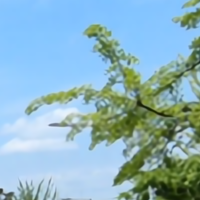}}
        \caption{ 28.01/0.1916}
    \end{subfigure}
    \hfill
    \begin{subfigure}[t]{0.15\textwidth}
        \centering
        \frame{\includegraphics[width=\textwidth]{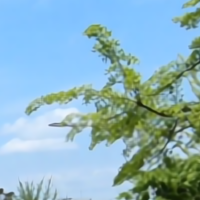}}
        \caption{28.51/0.1696}
    \end{subfigure}
    \hfill
    \begin{subfigure}[t]{0.15\textwidth}
        \centering
        \frame{\includegraphics[width=\textwidth]{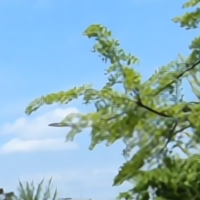}}
        \caption{28.65/0.1746}
    \end{subfigure}
    \hfill
    \begin{subfigure}[t]{0.15\textwidth}
        \centering
        \frame{\includegraphics[width=\textwidth]{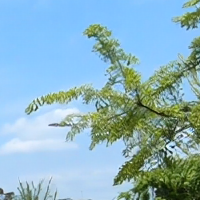}}
        \caption{26.04/0.1498}
    \end{subfigure}
    \hfill
    \begin{subfigure}[t]{0.15\textwidth}
        \centering
        \frame{\includegraphics[width=\textwidth]{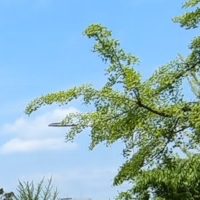}}
    \end{subfigure}

    \vspace{1mm} 

    \begin{subfigure}[t]{0.15\textwidth}
        \centering
        \frame{\includegraphics[width=\textwidth]{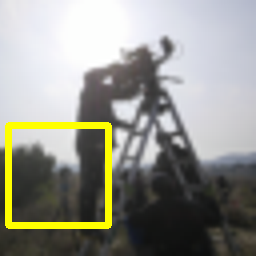}}
    \end{subfigure}
    \hfill
    \begin{subfigure}[t]{0.15\textwidth}
        \centering
        \frame{\includegraphics[width=\textwidth]{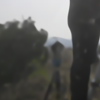}}
        \caption{34.07/0.2138}
    \end{subfigure}
    \hfill
    \begin{subfigure}[t]{0.15\textwidth}
        \centering
        \frame{\includegraphics[width=\textwidth]{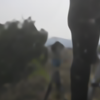}}
        \caption{34.06/0.2094}
    \end{subfigure}
    \hfill
    \begin{subfigure}[t]{0.15\textwidth}
        \centering
        \frame{\includegraphics[width=\textwidth]{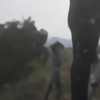}}
        \caption{34.11/0.2100}
    \end{subfigure}
    \hfill
    \begin{subfigure}[t]{0.15\textwidth}
        \centering
        \frame{\includegraphics[width=\textwidth]{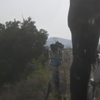}}
        \caption{32.38/0.1326}
    \end{subfigure}
    \hfill
    \begin{subfigure}[t]{0.15\textwidth}
        \centering
        \frame{\includegraphics[width=\textwidth]{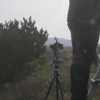}}
    \end{subfigure}

    \vspace{1mm} 

    \begin{subfigure}[t]{0.15\textwidth}
        \centering
        \frame{\includegraphics[width=\textwidth]{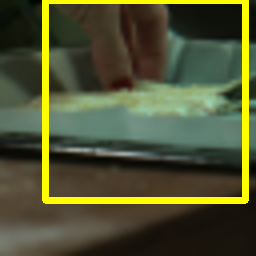}}
    \end{subfigure}
    \hfill
    \begin{subfigure}[t]{0.15\textwidth}
        \centering
        \frame{\includegraphics[width=\textwidth]{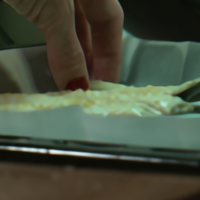}}
        \caption{38.44/0.1260}
    \end{subfigure}
    \hfill
    \begin{subfigure}[t]{0.15\textwidth}
        \centering
        \frame{\includegraphics[width=\textwidth]{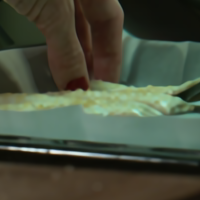}}
        \caption{38.54/0.1237}
    \end{subfigure}
    \hfill
    \begin{subfigure}[t]{0.15\textwidth}
        \centering
        \frame{\includegraphics[width=\textwidth]{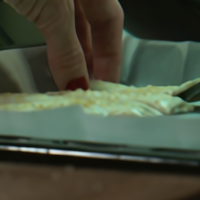}}
        \caption{38.94/0.1221}
    \end{subfigure}
    \hfill
    \begin{subfigure}[t]{0.15\textwidth}
        \centering
        \frame{\includegraphics[width=\textwidth]{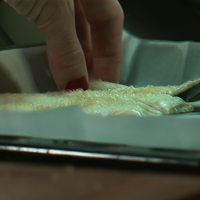}}
        \caption{36.72/0.0908}
    \end{subfigure}
    \hfill
    \begin{subfigure}[t]{0.15\textwidth}
        \centering
        \frame{\includegraphics[width=\textwidth]{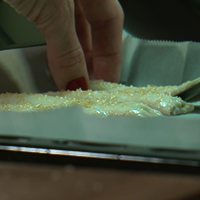}}
    \end{subfigure}

    \vspace{-3mm}

\caption{\textbf{Qualitative comparison with other baselines on public datasets (REDS4~\cite{Nah2019reds4}, Vimeo-90K-T~\cite{xue2019toflow}.} 
We show \textbf{PSNR/LPIPS} below each output frame.
PSNR does not align well with human perception and favor blurry results. LPIPS is a preferred metric that aligns better with human perception.
Compared to previous VSR approaches, our model can produce more realistic textures and more fine-grained details. 
}
\label{fig:qual_comparison}
\end{figure*}

%% file: figures/ablation.tex
\begin{figure*}[t!]
    \begin{minipage}{\textwidth}
    \begin{subfigure}[t]{0.15\textwidth}
        \centering
        \frame{\includegraphics[width=\textwidth]{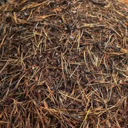}}
        \caption{Input}
    \end{subfigure}
    \hfill
    \begin{subfigure}[t]{0.15\textwidth}
        \centering
        \frame{\includegraphics[width=\textwidth]{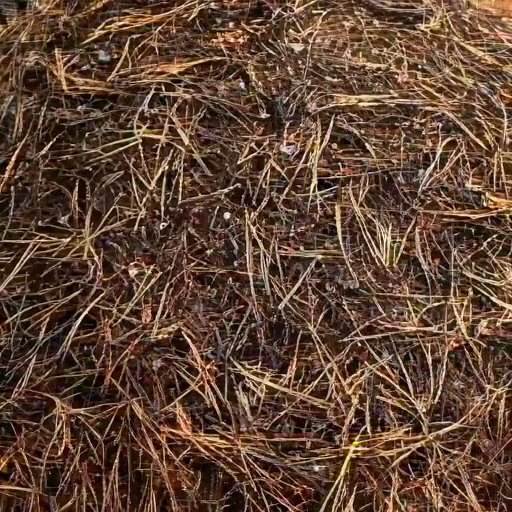}}
        \caption{+Temporal attention}
    \end{subfigure}
    \hfill
    \begin{subfigure}[t]{0.15\textwidth}
        \centering
        \frame{\includegraphics[width=\textwidth]{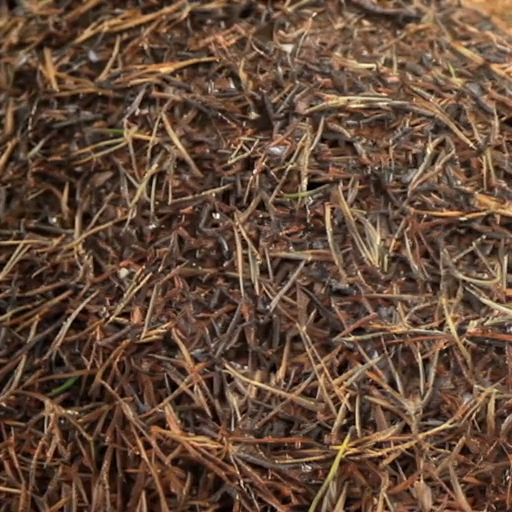}}
        \caption{+Feature propagation}
    \end{subfigure}
    \hfill
    \begin{subfigure}[t]{0.15\textwidth}
        \centering
        \frame{\includegraphics[width=\textwidth]{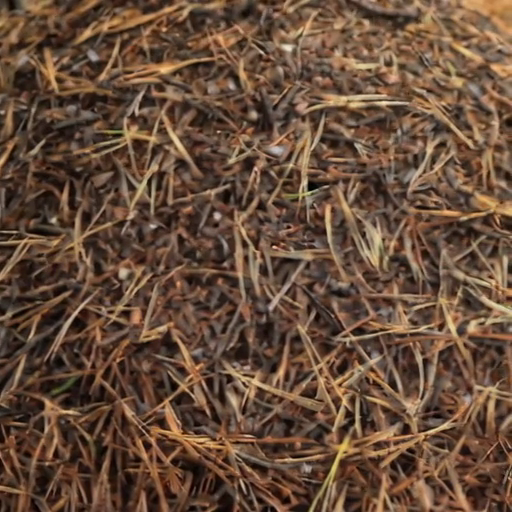}}
        \caption{+BlurPool}
    \end{subfigure}
    \hfill
    \begin{subfigure}[t]{0.15\textwidth}
        \centering
        \frame{\includegraphics[width=\textwidth]{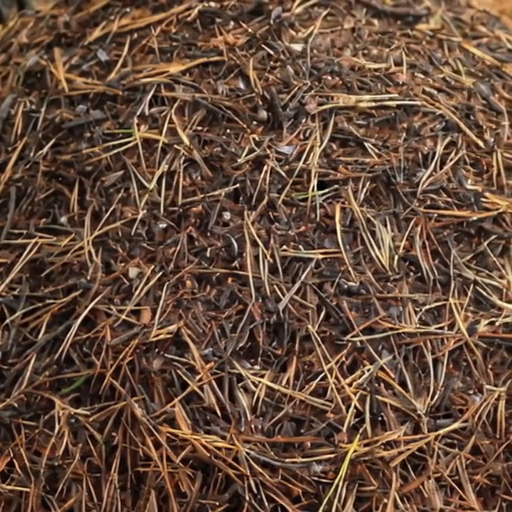}}
        \caption{+HF shuttle}
    \end{subfigure}
    \hfill
    \begin{subfigure}[t]{0.15\textwidth}
        \centering
        \frame{\includegraphics[width=\textwidth]{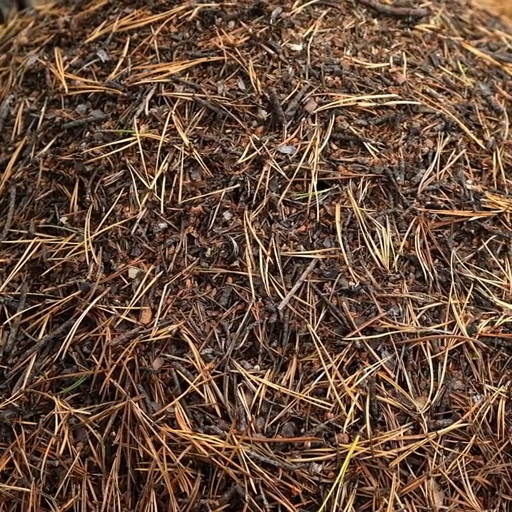}}
        \caption{GT}
    \end{subfigure}
    \end{minipage}

    \begin{subfigure}[t]{\textwidth}
        \centering
        \includegraphics[width=\textwidth]{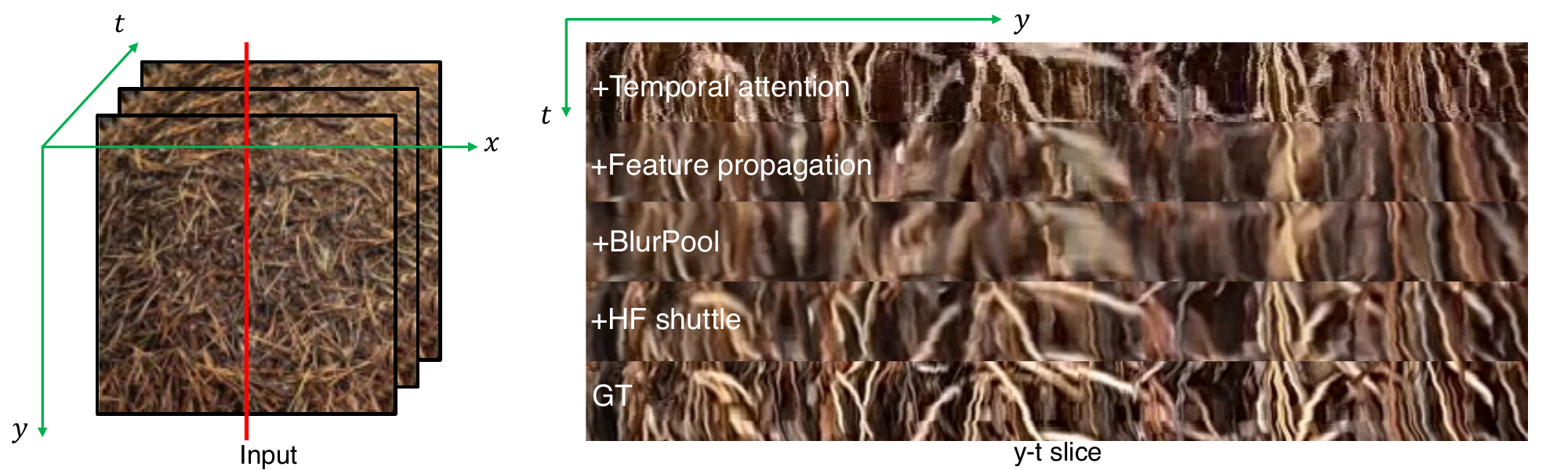}
        \label{fig:xt_slice_ablation}
    \end{subfigure}
    
    \vspace{-3mm}
\caption{\textbf{Ablation study.} 
Starting from the inflated GigaGAN (\textit{+Temporal attention} in the figure), we progressively add components to demonstrate its effectiveness.
With \textbf{temporal attention}, the local temporal consistency is improved compared to using image GigaGAN to upsample each frame independently.
The global temporal consistency improves with \textbf{feature propagation},  but aliasing still exists in the areas with high-frequency details (please refer to the videos in the \href{https://videogigagan.github.io/}{project website}). Also, the video results become more blurry.
By using the anti-aliasing blocks -- \textbf{BlurPool}, the aliasing issue is much better, but the video results become even more blurry.
Finally, with \textbf{HF shuttle}, we can bring the per-frame quality and high-frequency details back while preserving good temporal consistency.
}
\label{fig:ablation}
\end{figure*}

%% file: tables/ablation.tex
\begin{table}[h!]
  \centering
    \begin{tabular}{lrr}
    \toprule
    Model & \multicolumn{1}{l}{LPIPS$\downarrow$} & \multicolumn{1}{l}{$E^{ref}_{\mathrm{warp}} \downarrow (\times 10^{-3})$ } \\
    \midrule
    GigaGAN (base upsampler) &  0.2031     & 2.497 \\
    \midrule
    + Temporal attention &   0.2029   & 2.462  \\
    + Flow-guided propagation & $\boldsymbol{0.1551}$      & 2.187 \\
    + BlurPool &   0.1621    &  $\boldsymbol{2.152}$\\
    + High-freq shuttle &  $\underline{0.1582}$     & $\underline{2.177}$ \\
    \bottomrule
    \end{tabular}%
    \caption{\textbf{Ablation study.} We use LPIPS to evaluate per-frame quality and $E^{ref}_{\mathrm{warp}} \downarrow (\times 10^{-3})$ for temporal consistency. Starting from the image GigaGAN (upsampling each frame independently with the image upsampler), we progressively add components to demonstrate its effectiveness. The best number: \textbf{bold}. The second best number: \underline{underline}.}
  \label{tab:ablation}%
  \vspace{-5mm}
\end{table}%

%% file: figures/quality_vs_temp.tex
\begin{table}[h!]
    \scriptsize
        \begin{center}
            \begin{tabular}{lrrr}            
            \toprule
            Method & \multicolumn{1}{l}{LPIPS$\downarrow$} & \multicolumn{1}{l}{$E_{\mathrm{warp}} \downarrow (\times 10^{-3})$ } & \multicolumn{1}{l}{$E^{ref}_{\mathrm{warp}} \downarrow (\times 10^{-3})$ }\\
            \midrule            
            Bicubic & 0.3396 & \textbf{1.161} & 2.4232 \\
            \midrule
            EDVR~\cite{wang2019edvr} &  0.2097     &  1.521 & 2.1429\\
            MuCAN~\cite{li2020mucan} &   0.2162   &   1.562 & 2.1574\\
            BasicVSR~\cite{chan2021basicvsr} & 0.2023      &  1.371 & 2.1220\\
            IconVSR~\cite{chan2021basicvsr} &   0.1939    &  1.379 & 2.2119 \\
            TTVSR~\cite{liu2022learning}  &  0.1836     &  1.390 & \textbf{2.1178}\\
            BasicVSR++~\cite{chan2022basicvsrpp} & 0.1786 &  1.401 & 2.1206\\
            Ours &   \textbf{0.1582}  &  2.313 & 2.1773\\
            \midrule
            Ground truth   &   -  &  2.127 & 2.1272 \\
            \bottomrule
            \end{tabular}%
            \end{center}
    \caption{\textbf{Comparison of VideoGigaGAN and previous VSR approaches in terms of temporal consistency and per-frame quality.} The commonly used $E_{\mathrm{warp}}$ for temporal consistency favors more blurry results. The naive BICUBIC upsampling method achieves the lowest $E_{\mathrm{warp}}$. To address this issue, we propose to use the referenced warping error $E^{ref}_{\mathrm{warp}}$ for temporal consistency.}
    \label{tab:trade_off}
\end{table}


%% file: tables/main_baselines.tex
\begin{table*}[t!]
  \centering
  \tiny
    \begin{tabular}{l|ccc|ccc}
    \toprule
          & \multicolumn{3}{c|}{BI degradation} & \multicolumn{3}{c}{BD degradation} \\
    \midrule
          & REDS4~\cite{Nah2019reds4} & \multicolumn{1}{c}{Vimeo-90K-T~\cite{xue2019toflow}} & \multicolumn{1}{c|}{Vid4~\cite{liu2013vid4}} & \multicolumn{1}{c}{UMD10~\cite{PFNL}} & \multicolumn{1}{c}{Vimeo-90K-T} & \multicolumn{1}{c}{Vid4} \\
    \midrule
    TOFlow~\cite{xue2019toflow} & -/27.98 & \multicolumn{1}{c}{-/33.08} & \multicolumn{1}{c|}{-/25.89} & \multicolumn{1}{c}{-/36.26} & \multicolumn{1}{c}{-/34.62} & \multicolumn{1}{c}{-} \\
    RBPN~\cite{RBPN2019}  & -/30.09 & \multicolumn{1}{c}{-/37.07} & \multicolumn{1}{c|}{-/27.12} & \multicolumn{1}{c}{-/38.66} & \multicolumn{1}{c}{-/37.20} &  \multicolumn{1}{c}{-}\\
    PFNL~\cite{PFNL}  & -/29.63 & \multicolumn{1}{c}{-/36.14} & \multicolumn{1}{c|}{-/26.73} & \multicolumn{1}{c}{-/38.74} & \multicolumn{1}{c}{-} & \multicolumn{1}{c}{-/27.16} \\
    EDVR~\cite{wang2019edvr}  & 0.2097/31.05 & \multicolumn{1}{c}{-/37.61} & \multicolumn{1}{c|}{-/27.35} & \multicolumn{1}{c}{-/39.89} & \multicolumn{1}{c}{-/37.81} & \multicolumn{1}{c}{-/27.85} \\
    MuCAN~\cite{li2020mucan} & 0.2162/30.88 & \multicolumn{1}{c}{0.1523/37.32} & \multicolumn{1}{c|}{-} & \multicolumn{1}{c}{-} & \multicolumn{1}{c}{-} & \multicolumn{1}{c}{-} \\
    BasicVSR~\cite{chan2021basicvsr} & 0.2023/31.42 & \multicolumn{1}{c}{0.1616/37.18} & \multicolumn{1}{c|}{0.2812/27.24} & \multicolumn{1}{c}{0.1148/39.96} & \multicolumn{1}{c}{0.1551/37.53} & \multicolumn{1}{c}{0.2555/27.96} \\
    IconVSR~\cite{chan2021basicvsr} & 0.1939/31.67 & \multicolumn{1}{c}{0.1587/37.47} & \multicolumn{1}{c|}{0.2739/27.39} & \multicolumn{1}{c}{0.1152/40.03} & \multicolumn{1}{c}{0.1531/37.84} & \multicolumn{1}{c}{0.2462/28.04} \\
    TTVSR~\cite{liu2022learning} & 0.1836/32.12 & \multicolumn{1}{c}{-} & \multicolumn{1}{c|}{-} & \multicolumn{1}{c}{0.1112/40.41} & \multicolumn{1}{c}{0.1507/37.92} & \multicolumn{1}{c}{0.2381/28.40} \\  
    BasicVSR++~\cite{chan2022basicvsrpp}  & 0.1786/{32.39} & \multicolumn{1}{c}{0.1506/{37.79}} & \multicolumn{1}{c|}{0.2627/{27.79}} & \multicolumn{1}{c}{0.1131/{40.72}} & \multicolumn{1}{c}{0.1440/{38.21}} & \multicolumn{1}{c}{0.2390/{29.04}} \\
    RVRT~\cite{liang2022rvrt}  & 0.1727/\textbf{32.74} & \multicolumn{1}{c}{0.1502/\textbf{38.15} } & \multicolumn{1}{c}{0.2500/\textbf{27.99}} & \multicolumn{1}{c}{0.1100/\textbf{40.90}} & \multicolumn{1}{c}{0.1465/\textbf{38.59}} & \multicolumn{1}{c}{0.2219/\textbf{29.54}} \\
    \midrule
    Ours & \textbf{0.1582}/30.46 & \multicolumn{1}{c}{\textbf{0.1120}/35.97}      &  \multicolumn{1}{c|}{\textbf{0.1925}/26.78}      &  \multicolumn{1}{c}{\textbf{0.1060}/36.57}     &    \multicolumn{1}{c}{\textbf{0.1129}/35.30}   & \multicolumn{1}{c}{\textbf{0.1832}/27.04} \\
    \bottomrule
    \end{tabular}%
    \caption{\textbf{Quantitative comparisons of VideoGigaGAN and previous VSR approaches in terms of per-frame quality (LPIPS$\downarrow$/PSNR$\uparrow$) evaluated on multiple datasets. }
  We also report SSIM scores in the \href{https://videogigagan.github.io/assets/supp.pdf}{supplementary material}.
  }
  \label{tab:main_baselines}%
\end{table*}%

%% file: tables/modelsize_compute_comp.tex
\begin{table}[t!]
  \centering
  \small
    \begin{tabular}{lcc}
    \toprule
          & \#Params(M) & Runtime(ms) \\
    \midrule
    RBPN~\cite{RBPN2019} &   12.2    &  1507 \\
    EDVR~\cite{wang2019edvr}  &   20.6    & 378  \\
    BasicVSR~\cite{chan2021basicvsr}  &   6.3    & 63 \\
    IconVSR~\cite{chan2021basicvsr} &  8.7     & 70  \\
    BasicVSR++~\cite{chan2022basicvsrpp} &  7.3     &  77 \\
    \midrule
    Ours  &   369    &  295 \\
    \bottomrule
    \end{tabular}%
    \caption{\textbf{Comparison of model sizes and runtimes.}
  We compute runtimes per frame on $320\times180$ to $1280\times720$ on REDS4~\cite{Nah2019reds4}.
  Our VideoGigaGAN model has a competitive runtime with a larger model size.
  }
  \label{tab:modelsize_runtime}%
\end{table}%

%% file: 5_limitation.tex
\section{Limitations}
\label{sec:limitations}
Our model encounters challenges when processing extremely long videos (e.g., 200 frames or more). 
This difficulty arises from misguided feature propagation caused by inaccurate optical flow in such extended video sequences. 
Additionally, our model does not perform well in handling small objects, such as text and characters, as the information pertaining to these objects is significantly lost in the LR video input. 
Examples of these failure cases are illustrated in Fig.~\ref{fig:limitations}.

%% file: figures/limitations.tex
\begin{figure}[h]
\begin{center}
\centering



\begin{subfigure}[t]{0.45\textwidth}
        \frame{\includegraphics[trim=0 0 0 0, clip,width=\textwidth]{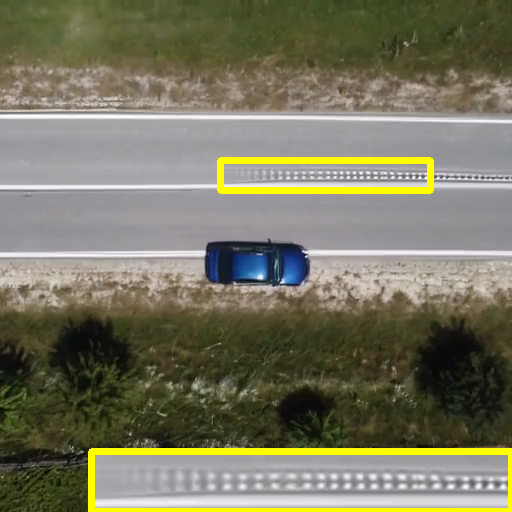}}
        \caption{(a) Extremely long video }
    \end{subfigure}
    \hfill
    \begin{subfigure}[t]{0.45\textwidth}
        \frame{\includegraphics[trim=0 0 0 0, clip,width=\textwidth]{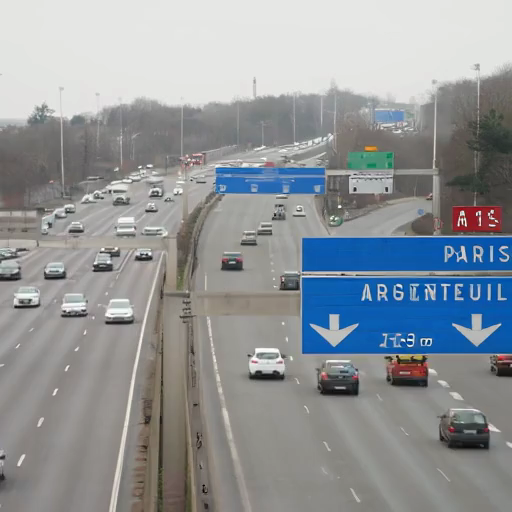}}
        \caption{ (b) Small objects}
\end{subfigure}

\caption{
\textbf{Limitations.} 
Our approach has some limitations. 
(a) When the video is \textbf{extremely long}, the feature propagation becomes inaccurate, which may introduce undesired artifacts like incorrect propagated patterns.
(b) Our model cannot handle well \textbf{small objects}, \eg small characters. 
}
\label{fig:limitations}
\end{center}
\end{figure}

%% file: 6_conclusion.tex
\section{Conclusions}
\label{sec:conclusions}
We present a novel generative VSR model, VideoGigaGAN, that can upsample input low-resolution videos to high-resolution videos with both high-frequency details and temporal consistency.
Previous VSR approaches often use regression-based networks and tend to generate blurry results. 
To this end, our VSR model built upon the powerful generative image upsampler -- GigaGAN. We identify several issues when applying GigaGAN to video super-resolution tasks including temporal flickering and aliased artifacts. To address these issues, we introduce new components to the GigaGAN architecture that can effectively improve both the temporal consistency and per-frame quality.
Our results demonstrate that VideoGigaGAN strike a balance in addressing the consistency-quality dilemma of VSR compared to previous methods.

%% file: tables/main_baselines_full.tex
\begin{table*}[h!]
  \centering
  \scriptsize
    \begin{tabular}{l|lrr|rrr}
    \toprule
          & \multicolumn{3}{c|}{BI degradation} & \multicolumn{3}{c}{BD degradation} \\
    \midrule
          & REDS4~\cite{Nah2019reds4} & \multicolumn{1}{l}{Vimeo-90K-T~\cite{xue2019toflow}} & \multicolumn{1}{l|}{Vid4~\cite{liu2013vid4}} & \multicolumn{1}{l}{UMD10~\cite{PFNL}} & \multicolumn{1}{l}{Vimeo-90K-T} & \multicolumn{1}{l}{Vid4} \\
    \midrule
    TOFlow~\cite{xue2019toflow} & -/27.98/0.7990 & \multicolumn{1}{l}{-/33.08/0.9054} & \multicolumn{1}{l|}{-/25.89/0.7651} & \multicolumn{1}{l}{-/36.26/0.9438} & \multicolumn{1}{l}{-/34.62/0.9212} & \multicolumn{1}{l}{-} \\
    RBPN~\cite{RBPN2019}  & -/30.09/0.8590 & \multicolumn{1}{l}{-/37.07/0.9435} & \multicolumn{1}{l|}{-/27.12/0.8180} & \multicolumn{1}{l}{-/38.66/0.9596} & \multicolumn{1}{l}{-/37.20/0.9458} &  \multicolumn{1}{l}{-}\\
    PFNL~\cite{PFNL}  & -/29.63/0.8502 & \multicolumn{1}{l}{-/36.14/0.9363} & \multicolumn{1}{l|}{-/26.73/0.8029} & \multicolumn{1}{l}{-/38.74/0.9627} & \multicolumn{1}{l}{-} & \multicolumn{1}{l}{-/27.16/0.8355} \\
    EDVR~\cite{wang2019edvr}  & 0.2097/31.05/0.8793 & \multicolumn{1}{l}{-/37.61/0.9489} & \multicolumn{1}{l|}{-/27.35/0.8264} & \multicolumn{1}{l}{-/39.89/0.9686} & \multicolumn{1}{l}{-/37.81/0.9523} & \multicolumn{1}{l}{-/27.85/0.8503} \\
    MuCAN~\cite{li2020mucan} & 0.2162/30.88/0.8750 & \multicolumn{1}{l}{0.1523/37.32/0.9465} & \multicolumn{1}{l|}{-} & \multicolumn{1}{l}{-} & \multicolumn{1}{l}{-} & \multicolumn{1}{l}{-} \\
    BasicVSR~\cite{chan2021basicvsr} & 0.2023/31.42/0.8909 & \multicolumn{1}{l}{0.1616/37.18/0.9450} & \multicolumn{1}{l|}{0.2812/27.24/0.8251} & \multicolumn{1}{l}{0.1148/39.96/0.9694} & \multicolumn{1}{l}{0.1551/37.53/0.9498} & \multicolumn{1}{l}{0.2555/27.96/0.8553} \\
    IconVSR~\cite{chan2021basicvsr} & 0.1939/31.67/0.8948 & \multicolumn{1}{l}{0.1587/37.47/0.9476} & \multicolumn{1}{l|}{0.2739/27.39/0.8279} & \multicolumn{1}{l}{0.1152/40.03/0.9694} & \multicolumn{1}{l}{0.1531/37.84/0.9524} & \multicolumn{1}{l}{0.2462/28.04/0.8570} \\
    TTVSR~\cite{liu2022learning} & 0.1836/32.12/0.9021 & \multicolumn{1}{l}{-} & \multicolumn{1}{l|}{-} & \multicolumn{1}{l}{0.1112/40.41/0.9712} & \multicolumn{1}{l}{0.1507/37.92/0.9526} & \multicolumn{1}{l}{0.2381/28.40/0.8643} \\  
    BasicVSR++~\cite{chan2022basicvsrpp}  & 0.1786/{32.39}/{0.9069} & \multicolumn{1}{l}{0.1506/{37.79}/{0.9500}} & \multicolumn{1}{l|}{0.2627/{27.79}/{0.8400}} & \multicolumn{1}{l}{0.1131/{40.72}/{0.9722}} & \multicolumn{1}{l}{0.1440/{38.21}/ {0.9550}} & \multicolumn{1}{l}{0.2390/ {29.04}/{0.8753}} \\
    RVRT~\cite{liang2022rvrt}  & 0.1727/\textbf{32.74}/\textbf{0.9113} & \multicolumn{1}{l}{0.1502/\textbf{38.15}/ \textbf{0.9527}} & \multicolumn{1}{l|}{0.2500/\textbf{27.99}/\textbf{0.8464}} & \multicolumn{1}{l}{0.1100/\textbf{40.90}/\textbf{0.9729}} & \multicolumn{1}{l}{0.1465/\textbf{38.59}/ \textbf{0.9576}} & \multicolumn{1}{l}{0.2219/\textbf{29.54}/\textbf{0.8811}} \\
    \midrule
    Ours &  \textbf{0.1582}/30.46/0.8718 & \multicolumn{1}{l}{ \textbf{0.1120}/35.97/0.9238}      &  \multicolumn{1}{l|}{ \textbf{0.1925}/26.78/0.8029}      &  \multicolumn{1}{l}{ \textbf{0.1060}/36.57/0.9521}     &    \multicolumn{1}{l}{ \textbf{0.1129}/35.30/0.9317}   & \multicolumn{1}{l}{ \textbf{0.1832}/27.04/0.8365} \\
    \bottomrule
    \end{tabular}%

  \caption{\textbf{Quantitative comparisons of VideoGigaGAN and previous VSR approaches.
  } 
  We report  LPIPS$\downarrow$/PSNR$\uparrow$/SSIM$\uparrow$.
  Similar to our conclusion in the main paper, our model achieves the lowest LPIPS error, showing a detail-rich result. 
  }
  \label{tab:main_baselines_supp}%
\end{table*}%

%% file: tables/gigagan_config.tex
\begin{table*}[t]
  \centering
  \caption{GigaGAN model configurations}
    \begin{tabular}{lc}
    \toprule
    $\mathbf{z}$ dimension & 512 \\
    $\mathbf{w}$ dimension & 512 \\
    Mapping network layers & 4 \\
    Activation & LeakyReLU \\
    $\mathcal{G}$ channel base & 32768 \\
    $\mathcal{G}$ channel max & 512 \\
    $\mathcal{G}$ \# of filters $N$ for adaptive kernel selection & [1, 1, 1, 1, 1, 2, 4, 8, 16, 16, 16, 16] \\
    $\mathcal{G}$ spatial self-attention resolutions & [8, 16] \\
    $\mathcal{G}$ temporal attention resolutions & [8, 16, 32, 64] \\
    $\mathcal{G}$ attention depth & [2, 2, 2, 1] \\
    $\mathcal{G}$ temporal attention window size & 1 \\
    $\mathcal{G}$ temporal convolution kernel size & 3 \\
    $\mathcal{G}$ \# synthesis block per resolution & [4, 4, 4, 4, 4, 4, 3] \\
    $\mathcal{G}$ \# downsampling blocks & 3 \\
    $\mathcal{D}$ channel base & 32768 \\
    $\mathcal{D}$ channel max & 512 \\
    $\mathcal{D}$ attention depth & [2, 2, 1] \\
    $\mathcal{D}$ attention resolutions & [8, 16] \\
    \midrule
    $\mathcal{G}$ model size & 369M \\
    $\mathcal{D}$ model size & 179M \\
    \bottomrule
    \end{tabular}%
  \label{tab:supp_model_config}%
\end{table*}%